\documentclass[letterpaper, 10 pt, conference]{ieeeconf}  

\IEEEoverridecommandlockouts                              

\usepackage{graphicx} 
\usepackage{cite}
\usepackage{amsmath,amssymb,amsfonts}
\usepackage{algorithmic}
\usepackage{textcomp}
\usepackage{xcolor}
\usepackage{threeparttable}
\usepackage{booktabs}
\usepackage{multirow}
\usepackage{makecell}
\usepackage{bbm}
\usepackage{multirow}
\usepackage{tabularx,booktabs}

\title{Using Physiological Measures, Gaze, and Facial Expressions to Model Human Trust in a Robot Partner}

\author{Haley N. Green and Tariq Iqbal
\thanks{Research reported in this paper is supported by the Air Force Office of Scientific Research Young Investigator Program (FA9550-24-1-0081).}
\thanks{The authors are with the School of Engineering and Applied Science, University of Virginia, Charlottesville, VA, USA.
        {\tt\small hng9vf@virginia.edu, tiqbal@virginia.edu.}}%
}

\begin{document}

\maketitle
\thispagestyle{empty}
\pagestyle{empty}

\begin{abstract}
With robots becoming increasingly prevalent in various domains, it has become crucial to equip them with tools to achieve greater fluency in interactions with humans. One of the promising areas for further exploration lies in human trust. A real-time, objective model of human trust could be used to maximize productivity, preserve safety, and mitigate failure. In this work, we attempt to use physiological measures, gaze, and facial expressions to model human trust in a robot partner. We are the first to design an in-person, human-robot supervisory interaction study to create a dedicated trust dataset. Using this dataset, we train machine learning algorithms to identify the objective measures that are most indicative of trust in a robot partner, advancing trust prediction in human-robot interactions. Our findings indicate that a combination of sensor modalities (blood volume pulse, electrodermal activity, skin temperature, and gaze) can enhance the accuracy of detecting human trust in a robot partner. Furthermore, the Extra Trees, Random Forest, and Decision Trees classifiers exhibit consistently better performance in measuring the person's trust in the robot partner. These results lay the groundwork for constructing a real-time trust model for human-robot interaction, which could foster more efficient interactions between humans and robots.

\end{abstract}


\section{Introduction}
Robots are being used as collaborative partners across various domains, from healthcare to education to manufacturing \cite{vaibhav, gizem,li22, elgarf, Gombolay2016RSS}. Within these dynamic environments, robots must be designed to adapt and facilitate more fluent and productive interactions \cite{mott, islam, hoffman2019evaluating,  Iqbal2021ICRA,Islam2021_RAL,yasar}. Researchers have worked to enhance the tools a robot can use to better perceive and interact with humans \cite{oliveira22, Birmingham, hedayati}. Trust has emerged as a critical metric for determining the human agent's willingness to interact with a robotic teammate. By establishing a balance of trust, the robot can maximize productivity while curating the best user experience that encourages future use. For example, in the context of home healthcare, the extent to which an older adult trusts a robot's ability to assist them will affect their interaction with the robot. If the person trusts the robot too little, it may not be able to effectively serve as a helpful assistant. Conversely, if the person trusts the robot too much, the robot may be unable to provide adequate support for a given task, which could put the user in danger. Ultimately, a real-time model of the human's trust in the robot could be used by the robot to better connect with users, set expectations, and strengthen working relationships.  


\begin{figure}
    \centering
    \includegraphics[width=\linewidth]{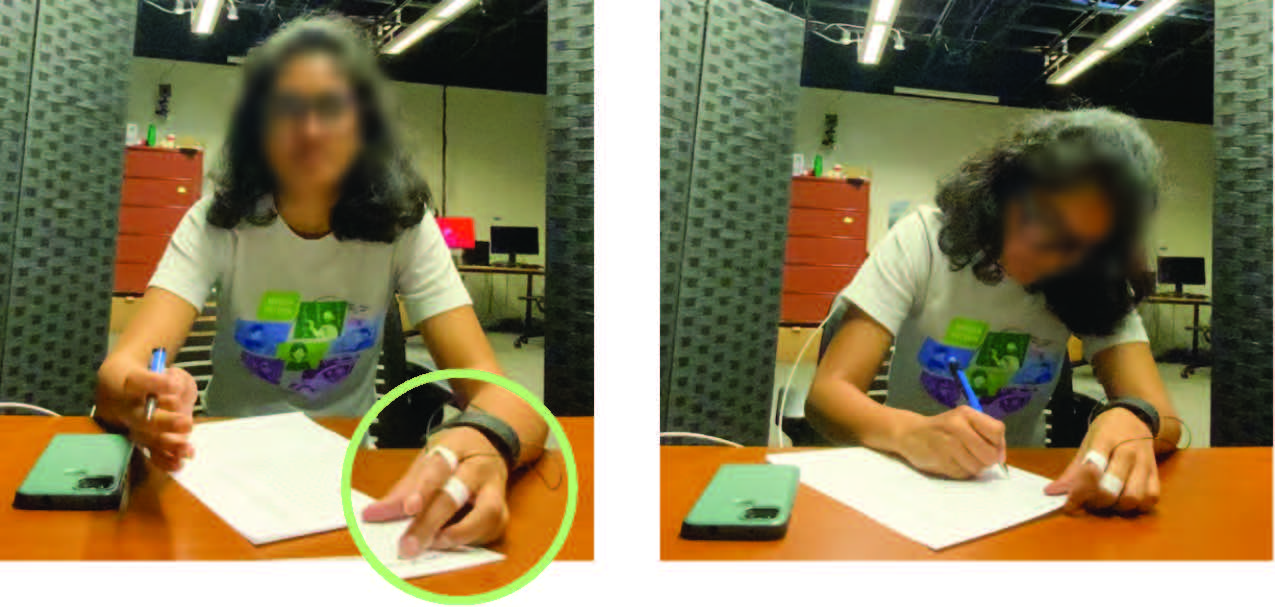}
    \vspace{-0.2in}
    \caption{Sample view of a participant supervising a robot (left) and completing their word search task (right). The Empatica E4 watch is worn on the non-dominant hand with EDA extender leads attached to the index and middle fingers, highlighted on the left.}
    \label{fig:participant_view}
    \vspace{-0.2in}
\end{figure}


Trust in human-robot interaction (HRI) has traditionally been measured using subjective questionnaires, which are effective in some contexts. However, this approach is not as suitable for real-time applications. Prior research has identified a relationship between trust and physiological response \cite{mitkidis2015building, potts2019trust, merrill2017trust}. However, these works often utilize more invasive and less readily accessible measures, which require sensors to be placed in restrictive areas such as the scalp or chest. Hence, to address this gap, we aim to foster greater connectivity in HRI by modeling trust in the multimodal, physiological space. Specifically, we aim to explore the use of non-invasive modalities for the objective measurement of human trust. {To ensure that our evaluation of trust is robust and that we are examining trust in our study design, we have designed our experiment based on the widely accepted definition of trust developed by Lee and See\cite{lee2004trust}: ``\textit{trust is the attitude that an agent will help achieve an individual’s goals in a situation characterized by uncertainty and vulnerability}''.}

The relationship between trust, affect, and physiological response has been extensively explored, though often separately. Trust has been linked to affective states such as stress and anxiety, which in turn are associated with physiological responses like elevated heart rate and changes in skin temperature. The connection suggests that physiological responses can serve as indicators of trust in HRI. To explore this relationship, we curated a novel human-robot trust dataset through a human-robot supervisory study that was designed to incite instances of trust and distrust in the human agent. We intentionally design our study to emulate a manufacturing environment. Within these environments, robots are used to reduce the mental and physical workload of the human workers. In particular, the human workers may be responsible for supervising the robots as they assemble various pieces. When there is trust in the robot partner, the human may spend less time directly observing its performance, thus allowing them to focus on alternative tasks. Moreover, we aim to consider trust as a scaled value. We believe that in representing trust on an interval scale, we can better encapsulate the dynamic nature of trust. This interval scale will enable us to develop robot behaviors that best calibrate the human's trust value.

For the physiological measures, we collect participant electrodermal activity (EDA), blood volume pulse (BVP), and skin temperature (TEMP) using a non-invasive sensor (i.e., Empatica E4 wristwatch \cite{empatica}). Additionally, for the gaze (GAZE) data collection, we consider gaze position in the users' view using the Pupil Invisible eye tracking glasses \cite{pupil}. Finally, we use action unit (AU) intensity to represent the facial expression of the human agent, captured using a high-definition external camera. The dataset, which we will open source for other researchers, contains a human's reported trust in a robot partner and the physiological measurements, gaze, and AUs associated with those trust levels.

To effectively predict real-time trust in HRI, we developed several trust models incorporating six widely used machine learning algorithms. In particular, we consider Random Forest (RF), Extra Trees (ET), Linear Discriminant Analysis (LDA), Logistic Regression (LR), Decision Tree (DT), and Support Vector Machine (SVM), which are commonly used to detect affective states. Moreover, we elected to incorporate these classifiers due to their accessibility and ease of incorporation. Unlike deep learning models, shallow learning models do not require large amounts of annotated training data. Ultimately, we aim to develop a model that is flexible enough to be effectively deployed in fast-paced manufacturing environments with little overhead.

Our results indicate that combining physiological data (BVP, EDA, and TEMP) and gaze data leads to improved accuracy in recognizing human trust in a robot partner. Specifically, we observed the top cross-validation accuracy of 97.5\% in the ET classifier that used a combination of EDA, BVP, TEMP, and GAZE. In addition, the ET, RF and DT classifiers show consistently improved accuracies across modality combinations compared to LDA, LR, and SVM. These findings provide the foundation for creating an objective model of trust in HRI, which could enable robots to be more accessible to users and foster more productive collaborations.

\section{Background}
\subsection{Trust in Human-Human Interactions}
Trust is a pervasive factor in a person's decision-making process, facilitating connections between individuals while fostering productive collaborations \cite{luhmann, kramer1999trust,gambetta2000can}. A person's trust is shaped by a variety of both internal and external factors. Specifically, external factors such as task risk, situational complexity, and shared goals can influence a person's inclination to trust in a partner \cite{molm2000risk, dayan2010impact, chow2008social}. Conversely, trust levels can also be impacted by internal factors, unique to the individual (i.e., past experiences, emotion, and personality) \cite{fareri2012effects, lount2010impact, mooradian2006trusts}. Because both internal and external factors can be individualized and change over time, trust also changes depending on the person, environment, and situation. Previous work has explored the dynamic nature of trust and how it is affected by both internal and external changes \cite{vanneste2014trust, fulmer2013trust, jones1998experience, kim2009repair}.

There has been extensive research on how trust develops between humans. However, while humans are equipped with the skills to innately gauge and calibrate trust, robots must be designed to model the construct. Trust is essential to any successful collaboration, regardless of the anthropomorphism of the interaction partner. Thus, we aim to use the fundamental principles of trust in a human-human interaction to design a study to explore trust in a human-robot setting.

\begin{figure}
    \centering
    \includegraphics[width=\linewidth]{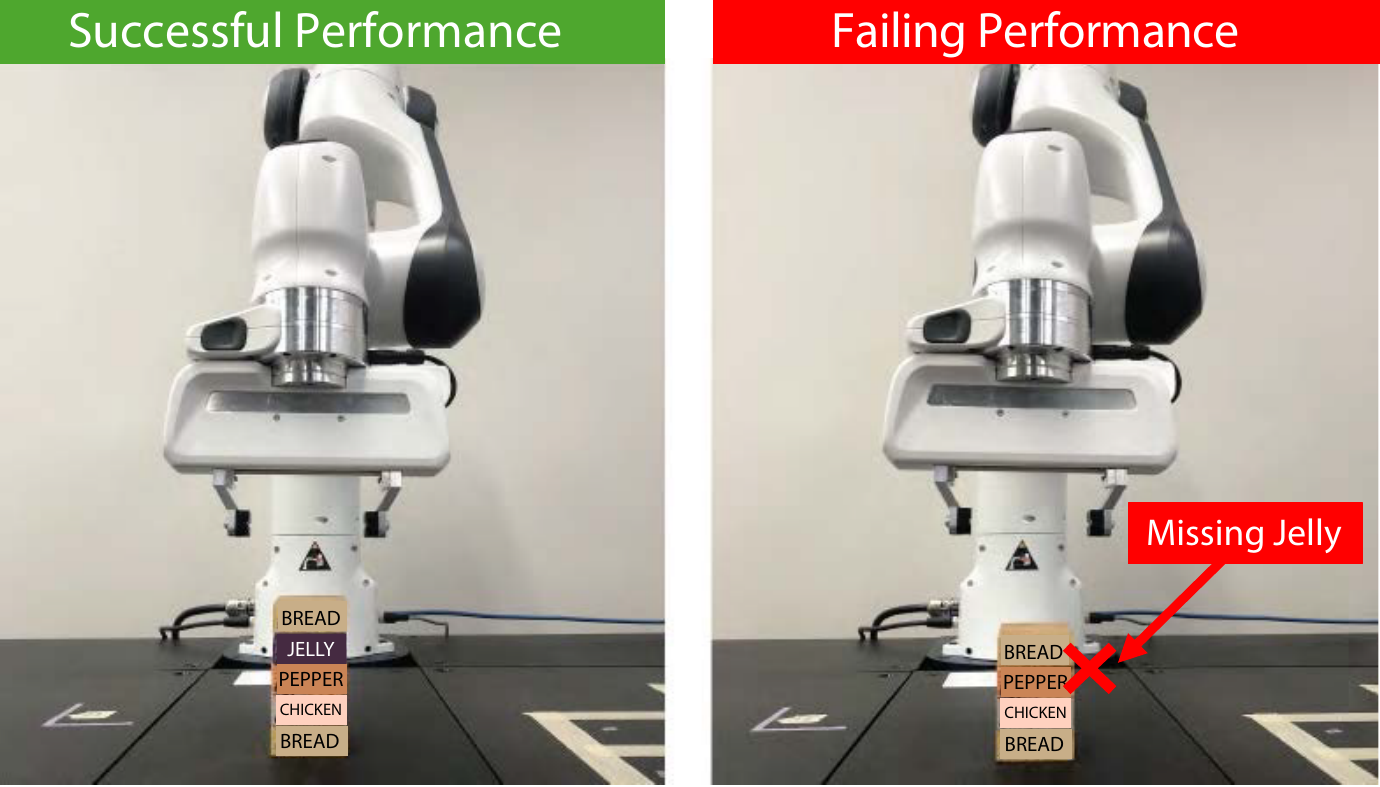}
    \caption{Sample view of the successful robot's performance (left) and the failing robot's performance (right). In the failure case, the robot stores the final ingredient instead of stacking it in the sandwich.}
    \label{fig:sandwich}
    \vspace{-0.1in}
\end{figure}

\subsection{Trust in Human-Robot Interactions}
Recent work in HRI has explored how humans trust in their robot partners and how that trust can be modeled to encourage more productive interactions \cite{herse, xu, hancock, kok2020trust, azevedo2021unified, chen2020trust}. Chen et al. used a partially observable Markov decision process to model trust in a robot decision-making task \cite{chen2020trust}. Prior work has also been conducted on how robots embody trustworthiness, respond in cases of failure to recover trust, and perform to encourage trust \cite{bryant, greenHRI, robinette2017}. For example, Natarajan and Gombolay studied the effects of anthropomorphism in a robot partner on trust \cite{natarajan}. Green et al. examined the effects of humor on maintaining trust and mitigating failure in HRI \cite{green}. Robinette et al. explored the effects of a robot's performance on human trust in time-critical scenarios \cite{robinette}. Prior work has also explored how robots can utilize human-like communication strategies to build trust in an interaction \cite{han, williams, martelaro, clair}.

Whether robots can achieve success in human-occupied spaces largely depends on the extent to which humans are willing to trust them. Robots must primarily be designed to succeed as assistive partners, aids, and educators. However, it is crucial that they are also equipped with the tools to ensure user safety, personalize user experience, and, in some cases, recover from task failure. Current methods for identifying trust do not incorporate objective measures, which limits the potential for using trust in real-time robot decision-making. Thus, we aim to address this research gap in this work. Having an adaptable, real-time measure of human trust in a robot partner could enable robots to determine the most appropriate strategies to maximize the productivity, safety, and experience of an interaction.

\subsection{Trust and Physiological Response}
Trust is intricately connected to various affective states, such as stress, anxiety, and anger \cite{schule,myers2016influence,dunn2005feeling}. Previous literature has identified that that affective states, including stress and anxiety, can impact human physiological responses. Specifically, increased stress and anxiety have been associated with elevated blood pressure, heightened cortisol levels, increased heart rate, and even shortness of breath \cite{chu2024physiology}. Additionally, stress has been shown to affect heart rate \cite{kim2018stress}, skin temperature \cite{herborn2015skin}, and electrodermal activity \cite{reinhardt2012salivary}. Akalin et al. explored how physiological data could be used to represent human perceptions of safety in a robot partner \cite{akalin2022you}. In social HRI settings, Ahmad and Alzahrani examined physiological measures and gaze as predictors of trust in an interactive game \cite{ahmad2023crucial}. 

Prior literature has established a relationship between trust and affective states, as well as a relationship between affective states and physiology. In this work, rather than attempting to individually interpret each internal and external affective state that makes up trust, we aim to directly relate trust to physiological response, gaze, and facial expression. 


\section{Method}
\subsection{Experimental Scenario} 
{We have structured our experiment around the widely recognized definition of trust by Lee and See \cite{lee2004trust} and utilize previously validated subjective survey measures to compare against our objective trust measures, providing a comprehensive assessment of trust in our study.} In order to develop a scenario that contains instances of trust and distrust, we modify the robot's performance for the different groups to alter the belief that the robot will successfully aid in achieving the team's goal. We also create a sense of vulnerability and uncertainty by establishing that the participant's compensation will depend on the team's performance. Thus, creating an opportunity for users to experience variations in their trust.

\subsubsection{Primary Task - Sandwich Assembly}
For the primary task, participants were asked to monitor the performance of a robot as it made sandwiches over the course of four tasks. At the start of the experiment, the participant identified a set of four sandwiches for the robot to make during the study. Then, the participant would supervise the robot as it stacked ingredients to form the four different sandwiches. Participants were instructed to report any observed errors in the robot's performance as soon as they were made in order to succeed in the primary task.


During the study, participants were randomly placed into two groups with conditions that could foster a variation in trust. Specifically, in the \textit{successful condition}, the robot successfully made all four sandwiches. In the \textit{failing condition}, the robot failed to make the last two sandwiches correctly by misplacing an ingredient in the storage bin (Fig. \ref{fig:sandwich}).

\begin{table}
  \centering
  \caption{The Sandwiches and their Ingredients.}
  \label{tab:sandwiches}
  \begin{tabular}{c|ccc}
  \toprule
    Sandwich&Meat&Vegetable&Condiment\\
    \hline
    A & Ham & Lettuce & Mustard\\
    B & Bacon & Tomato & Mayo\\
    C & Chicken & Pepper & Jelly\\
    D & Ham & Tomato & Jelly \\
    E & Bacon & Pepper & Mustard \\
    F & Chicken & Lettuce & Mayo \\
    \bottomrule
\end{tabular}
\vspace{-0.2in}
\end{table}

\subsubsection{Secondary Task - Word Searches}
 For the secondary task, we created a situation where participants had to decide whether to prioritize their own task or supervise the robot's performance. The secondary task consisted of several word searches to be completed as the robot was assembling the different sandwiches. The keywords used in the word searches were food-related in order to align with the context of the study. This secondary task was included based on previous research suggesting that increasing the task complexity can impact the level of trust in the agent \cite{bailey2007automation, zeng2006spontaneous}. 

To provide additional incentive to successfully complete both the primary (robot supervision) and secondary (word search) tasks, participants were informed that they could receive an additional \$5 (the base compensation was \$10) based on their team’s performance. The bonus was dependent on their supervisory abilities as well as the number of word searches completed. However, to ensure that all participants were fairly compensated for their time, we compensated all participants the full \$15, regardless of performance. All participants were notified of the deception at the conclusion of the study during the debriefing. All components of our study (including the deception component) were approved by the Institutional Review Board.

\subsection{The Robot}
In this study, the Franka Emika ``Panda'' \cite{panda} robot was used for the task of grasping, sorting, and stacking various objects. The Panda arm's movements were programmed in C++, with goal positions collected through the robot's impedance control. Joint positions were recorded at each location and used to ensure that the robot's performance was consistent across conditions. In the sandwich assembly task, the robot started from an initial position and worked iteratively through each ingredient. In the successful condition, all five ingredients were stacked, whereas in the failing condition, the robot incorrectly dropped the fourth ingredient into storage before stacking the final bread slice.

\subsection{Participants}
A total of 30 adults participated in the study (33.3\% female ($n = 10$), 63.3\% male ($n = 19$), and 3.3\% non-binary ($n = 1$)). The mean age of participants was $25.6$ years, $SD = 4.62$. All participants were required to be English speakers and at least 18 years of age. Participants also recorded their previous experience with robots on a Likert scale from ``no experience'' (1) to ``expert-level experience'' (5). The mean participant's previous robot experience was $2.83$, $SD = 1.09$. We had an equal number of participants for the two performance conditions.

\section{Measurement}
\subsection{Objective Measures}
\subsubsection{Physiological Measures}
We collected EDA, BVP, and TEMP data using the Empatica E4 smartwatch \cite{empatica}. The watch was fastened to the participant's non-dominant wrist to minimize any changes in physiological response while participants were writing. To obtain a more accurate reading of the EDA data, we also incorporated a set of sensor extender leads, which were attached to the inside base of the participant's index and middle fingers (Fig. \ref{fig:participant_view}). For the physiological data processing, we took participant EDA ($\mu$S), BVP ($\mu$V), and TEMP (C$^{\circ}$) for the session. Each sample contained the data timestamp and sampling frequency. During the study, we tagged the start of each four-minute task to mark the start of the event period of interest.

\subsubsection{Gaze}
Participant gaze data was collected using the Pupil Invisible eye tracking glasses \cite{pupil}. The glasses utilize a deep learning powered eye-tracking system to remove the need for time-consuming calibration. Thus, allowing the glasses to be easily used to collect accurate, real-time gaze data. The glasses are equipped with two eye cameras, a scene camera, a microphone, and an IMU. In our model, we use the x- and y-coordinates of gaze from the users' point-of-view.

\subsubsection{Facial Expression}
To examine participant facial expressions, we collected video data of the participant's performance during the task. The camera was positioned in front of the participant (Fig. \ref{fig:participant_view}) and was set to record the participant's secondary task (i.e., word search) performance and facial expressions. There are 46 Main Action Units in the face that can be used to determine emotion \cite{ekman1978facial}. However, we only consider a subset of 17 face-centric AUs for our model. 

\subsection{Subjective Trust Measures}
{We incorporated the Muir trust survey \cite{muir1996trust}, a well-established measure in HRI studies \cite{desai,chen2018planning,washburn2020robot}, on a Likert scale from ``strong distrust'' (1) to ``strong trust'' (5) to measure the subjective trust in the robot and generate the labels for our model.} After each task, participants were asked in a survey to self-report their trust in the robot. For the dataset, this label was assigned to the entirety of the session. Meaning, the reported trust for task one was assigned as a label for all of the physiological data generated in task one.

\section{Data Processing and Evaluation}
\subsection{Feature Extraction}
Each task was approximately 4 minutes in length, and the robot would stack or sort 11 blocks (ingredients) during this time. To account for variations in physiology, gaze, or facial expressions over the robot's performance, we selected a window length of 30 seconds with a 10-second slide to cover each of the blocks. A 30-second window provided an adequate time frame for capturing meaningful multimodal data, while a 10-second slide enabled us to sufficiently account for sudden fluctuations in physiology, gaze, and facial expressions. For the features, we considered the mean and standard deviation for the various signals. 

\subsection{Physiological Data Processing}
For the physiological measures, we collected data for BVP, EDA, and TEMP. The BVP signal is obtained from the photoplethysmography sensor and has a fixed sampling rate of 64 Hz. The BVP usually falls within the range of [-500, 500]. Both the EDA and TEMP signals were collected at a frequency of 4 Hz. To accommodate variations in individual physiological signals, we used the baseline session to normalize the physiological measures. Specifically, for each participant, we take their average EDA, BVP, and TEMP during the baseline session and subtract the value from their subsequent task session data. Thus, we are able to account for the physiological differences between participants.

\subsubsection{Baseline Session}
{Before interacting with the robot, participants were required to complete a baseline measurement session. This session was used to collect individual participant baseline physiological measures, which were later used to normalize the data for analysis. Similarly to the tasks, the baseline session was conducted over a span of four minutes. During the baseline session, participants were asked to sit still with their arms resting comfortably on the table. Participant data was normalized using $normalized = (a-b)$, where $a$ is the participant-specific mean value during the baseline measurement and $b$ is the measured data point.}

\subsection{Gaze Data Processing}
For the gaze data, we isolated the px and py values for the event epochs of interest. The data is collected at a rate of 200 Hz. We segmented the events into 30-second windows with 10-second slides to account for changes during the tasks. 

\subsection{Facial Expression Data Processing}
To incorporate facial expressions into our model, we collected video data of the participant's performance during each task. The videos were cropped to include only the relevant epoch windows and the sampling rate was about 30 frames per second. The resulting four-minute videos were then matched with the start/end tag on the Empatica E4 watch. We used OpenFace \cite{openface2} to generate action unit (AU) predictions for each of the videos. Next, the average task-specific AU intensities for each participant were calculated for each slide and window. Similarly to the physiological and gaze data, we modeled the AU data in 30-second windows with a 10-second slide to account for any changes throughout the tasks. The AU intensity values were continuous and on a scale from 0 (not present) to 5 (maximum intensity).

\subsection{Classifiers}
We identified six machine learning algorithms, each with varying strengths, to train on the dataset to identify which physiological responses are indicative of trust. Specifically, we considered Random Forest (RF), Extra Trees (ET), Linear Discriminant Analysis (LDA), Logistic Regression (LR), Decision Tree (DT), and Support Vector Machine (SVM), which are widely used in the literature for detecting various affective states. {Each of these classifiers has its merits, and we wanted to compare their performance to explore which one was the most effective for measuring trust.}

\subsubsection{Support Vector Machine} SVMs are common in image recognition and natural language processing applications. They are robust to noise and outliers, versatile, memory efficient, and consist of flexible kernel functions.

\subsubsection{Decision Trees} These are useful for data mining, classification, and regression applications. They are relatively easy to interpret, able to handle both categorical and numerical data, generally non-parametric by nature, and able to handle interactions and non-linear relationships.

\subsubsection{Random Forest} These classifiers are useful in regression and classification tasks. RF classifiers are resistant to overfitting and can handle non-linear relationships.

\subsubsection{Linear Discriminant Analysis} LDA models are utilized in dimensionality reduction and classification tasks. These models are relatively easy to interpret, robust to outliers, and able to avoid overfitting.

\subsubsection{Logistic Regression} These models are typically used in binary classification problems; however, they are equipped to handle complex datasets and are robust to noise. LR models also tend to have a lower computational cost.

\subsubsection{Extra Trees} This model is similar to RF and is common in classification and regression applications. ET classifiers are able to perform well even in cases of small sample sizes and have high performance on noisy data.


\subsection{Evaluation}
We performed 10-fold cross-validation to evaluate the performance of our models. We elected to use cross-validation as it provides a more reliable estimate of model performance than a single train-test split. Moreover, cross-validation allows for more efficient use of data since all data is used for training and testing. Finally, cross-validation is both resilient to overfitting and more generalizable to new data. Our aim is to develop a real-time model of human trust, thus, our analysis needs to be able to adapt to incorporate future datasets. To evaluate, we calculated the average cross-validation accuracy for the different modalities and classifiers.






    




\begin{table}[t]
\centering
\caption{Mean cross-validation accuracy (\%) (higher is better).}
\label{tab:mean_accuracy}
\resizebox{\columnwidth}{!}{%
\begin{tabular}{l|cccccc}
\toprule
Modality&RF&ET&LDA&LR&DT&SVM\\
\midrule
BVP & 52.6 & 52.6 & 55.3 & 55.1 & 46.2 & 55.0 \\
AU & 57.7 & 57.5 & 54.6 & 55.2 & 43.6 & 55.8 \\
TEMP & 64.8 & 64.1 & 52.7 & 52.6 & 62.8 & 56.1 \\
EDA & 67.3 & 67.9 & 53.4 & 62.1 & 62.5 & 64.2 \\
GAZE & 72.6 & \textbf{72.8} & 57.0 & 57.1 & 63.8 & 64.7 \\
\midrule
BVP-AU & 59.9 & 58.1 & 55.1 & 55.4 & 46.9 & 56.5 \\
TEMP-AU & 60.1 & 58.7 & 55.7 & 56.3 & 54.3 & 56.3 \\
GAZE-AU & 60.4 & 61.1 & 57.3 & 57.1 & 55.4 & 57.3 \\
EDA-AU & 64.7 & 61.4 & 56.7 & 60.4 & 56.7 & 55.9 \\
EDA-BVP & 76.5 & 77.0 & 56.3 & 61.7 & 68.8 & 63.4 \\
BVP-GAZE & 76.2 & 77.2 & 57.9 & 57.3 & 67.8 & 64.2 \\
BVP-TEMP & 76.8 & 78.0 & 53.7 & 54.5 & 70.9 & 61.0 \\
EDA-GAZE & 88.2 & 90.4 & 56.4 & 57.8 & 81.5 & 74.0 \\
TEMP-GAZE & 88.6 & 91.3 & 60.0 & 57.0 & 81.0 & 73.3 \\
EDA-TEMP & 89.8 & \textbf{92.5} & 59.6 & 62.5 & 85.4 & 71.8 \\
\midrule
BVP-TEMP-AU & 62.6 & 59.4 & 55.2 & 56.3 & 58.7 & 56.9 \\
BVP-GAZE-AU & 64.4 & 61.4 & 57.1 & 57.2 & 60.3 & 57.5 \\
EDA-BVP-AU & 66.5 & 63.7 & 57.3 & 59.1 & 61.9 & 57.7 \\
TEMP-GAZE-AU & 68.7 & 65.0 & 59.4 & 57.1 & 71.8 & 60.1 \\
EDA-GAZE-AU & 70.3 & 67.8 & 57.8 & 57.9 & 72.8 & 59.9 \\
EDA-TEMP-AU & 69.9 & 65.3 & 60.9 & 62.6 & 79.5 & 60.1 \\
BVP-TEMP-GAZE & 89.7 & 91.5 & 59.0 & 57.2 & 81.7 & 73.3 \\
EDA-BVP-GAZE & 89.0 & 90.6 & 56.9 & 57.8 & 80.9 & 73.0 \\
EDA-BVP-TEMP & 90.9 & 93.9 & 59.6 & 59.7 & 86.3 & 69.9 \\
EDA-TEMP-GAZE & 95.1 & \textbf{96.8} & 63.2 & 57.8 & 89.0 & 82.1 \\
\midrule
BVP-TEMP-GAZE-AU & 71.7 & 65.8 & 59.5 & 57.3 & 75.6 & 60.8 \\
EDA-BVP-GAZE-AU & 73.0 & 70.6 & 57.6 & 57.3 & 75.9 & 60.0 \\
EDA-BVP-TEMP-AU & 72.6 & 66.6 & 60.3 & 61.4 & 81.3 & 60.6 \\
EDA-TEMP-GAZE-AU & 77.9 & 74.8 & 61.5 & 58.7 & 85.1 & 64.0 \\
EDA-BVP-TEMP-GAZE & 94.9 & \textbf{97.5} & 62.9 & 57.2 & 90.0 & 79.2 \\
\midrule
EDA-BVP-TEMP-GAZE-AU & 79.5 & 75.0 & 61.8 & 57.5 & \textbf{86.8} & 64.6 \\
\bottomrule
\end{tabular}%
}
\vspace{-0.1in}
\end{table}

\section{Results and Discussion}
\subsection{Model Accuracy}
We present the average cross-validation accuracy for the various sensor modalities and classifiers in Table \ref{tab:mean_accuracy}. 
The results suggest that the ET classifier that used the EDA, BVP, TEMP, and GAZE data achieved the highest cross-validation accuracy at 97.5\%. The next highest accuracy was observed at 96.8\% with the ET classifier that used EDA, TEMP, and GAZE data. The inclusion of multiple features enables us to incorporate a diverse set of data into the trust model. The multitude of features can more accurately portray the human's trust while reducing the opportunity for overfitting due to a lack of robust feature options. The ET classifier further improves the accuracy of the model by mitigating the noise that is present due to the nature of the physiological features used. We posit that the combination of EDA, BVP, TEMP, and GAZE provides sufficient, complimentary information to the model without introducing too much redundancy. 

The lowest cross-validation accuracy was 43.6\% for the DT classifier that utilized only the AU data. It could be that by introducing the 17 AUs associated with the facial expression component, there may be too much redundant information for the model to generate more accurate predictions. Additionally, video data is highly dependent on consistent lighting conditions, clear camera angles, and unobstructed face poses. Any variations in the conditions in which the data is collected can add noise to the AU features. 



More generally, the cross-validation accuracies for the ET, DT, and RF classifiers tended to be higher than those for LDA, LR, and SVM. It could be that these classifiers are more robust to the noise typically associated with physiological data (e.g., environmental, measurement, biological, and motion). In our experiment, further measurement noise can be caused by unexpected participant motion, which impacts the physiological signal. ET and RF classifiers exhibit higher accuracies in scenarios with multiple modalities. In contrast, LDA and LR are simpler models that may have limitations in handling more complex datasets. Thus, when representing a real-world dataset such as this, which is inherently more complex and noisy due to the nature of the physiological data being collected, the ET, DT, and RF classifiers yield more accurate predictions compared to the LDA, LR, and SVM classifiers.

\subsection{Effect of Modality}
We observed that for the modality pairings, the AUs tend to lower the accuracy of the model's performance. Generally, the pairings of EDA, TEMP, and GAZE significantly enhance the accuracy of the model's predictions. For all the multimodality combinations, some combination of EDA and TEMP results in the highest accuracy. Moreover, GAZE as a single modality is more accurate than the other single modalities. When combined with TEMP, it tends to further increase the model's accuracy. For the single-modality models, those incorporating the BVP features tend to lower the accuracy of the predictions. Additionally, we observed that the models using both BVP and AUs resulted in the least accurate predictions across all modality pairings.

We posit that EDA, TEMP, and GAZE may be the driving predictive features when it comes to predicting human trust in a robot partner. For physiology, it could be that EDA and TEMP are more closely linked to emotional arousal, as indicated in prior work\cite{boucsein2012electrodermal,wilke2007short}. For example, EDA is a measure of the electrical conductance of the skin and is associated with sweat gland activity. EDA has previously been used to indicate emotional arousal stress levels \cite{boucsein2012electrodermal}. Similarly, TEMP is influenced by changes in blood flow and can be indicative of emotional arousal and stress \cite{wilke2007short}. Therefore, EDA and TEMP could be more insightful in detecting a person's trust. For GAZE, the nature of this task is supervisory, so we theorize that GAZE plays an important role in predicting human trust. When participants trust the robot, they may spend more time completing the secondary task instead of supervising the robot's performance. 

The reduced accuracy observed in the BVP and AU features could be the result of increased noise within the data streams. BVP data, in particular, is susceptible to noise interference. Additionally, facial expressions are highly dependent on the camera angle and lighting conditions. Since there were extended periods where participants were looking down to complete their secondary task, it could be that AUs collected from video data are less reliable indicators for this nature of supervisory tasks. Furthermore, our use of a large subset of AUs for the facial expression features may have added complexity to the model without improving the accuracy. Further reduction of the AUs to focus on a more concise and applicable subset may yield more accurate results, which we aim to explore in the future.


\section{Overall Discussion}
Our results indicate that physiological data can be usedF to model human trust in a robot partner. A combination of EDA, TEMP, and GAZE data yields higher accuracies for recognizing trust in a robot partner. Our findings suggest that adding BVP to the the model that uses EDA, TEMP, and GAZE can further enhance the accuracy. Furthermore, the ET, RF, and DT classifiers offer higher accuracies.

In this work, we examine trust on a 5-point Likert scale (``strong distrust'' to strong trust'') over a binary scale (``trust'' or ``distrust'' because humans do not necessarily experience trust in purely binary terms. Notably, the distribution of trust labels within the dataset is not even, which is reflective of trust in real-world scenarios. In the future, we would like to explore the model against more polarized ratings of trust. Ultimately, we recognize that people experience trust along a spectrum, and we believe that a 5-point scale is a suitable initial approach to capture variations in trust.

Our dataset and evaluation offer several advantages to the HRI community. Particularly, a real-time model of trust could be used to enable robots to facilitate more meaningful interactions with humans. Physiological measures, including BVP, EDA, and TEMP, along with gaze measures such as px and py and facial AUs, can be non-invasively collected through readily available tools such as smartwatches, gaze tracking glasses, and camera systems. {Given that physiological data is already collected by commercial smart health devices and voluntarily shared across various health-monitoring applications \cite{lui2022apple,gan2016overview,evans2017myfitnesspal}, it is reasonable to assume that users may also be willing to share this data with a robotic partner to enhance their interactions, provided it is used ethically and transparently.} Furthermore, these features can be collected and personalized to the user without lengthy calibration periods. In addition, by incorporating participant baseline physiology into the dataset, the model has the potential to be personalized to the individual. This will enable the proposed approach to be more accessible to users. Since the model demonstrated consistent accuracy across a variety of modality combinations, the features employed could be tailored to accommodate user preferences while maintaining consistently reliable trust predictions. Moreover, by considering trust as a scale ranging from ``strong distrust'' to ``strong trust,'' we have created opportunities for the model to be used to make more incremental adjustments. For example, if a human has ``strong distrust'' in the robot partner, the robot may need to take more drastic measures to increase trust than if the human were to simply ``distrust'' the robot. Finally, by using shallow machine learning models, we are able to generate predictions relatively quickly. These models are computationally inexpensive, which enables them to be used in real-time predictions deployed on robotic systems. 

\section{Conclusion}
{In this work, we used physiological measures, gaze, and facial expressions to generate a multimodal trust model. To achieve this, we produce a novel dataset of human trust in a robot partner and analyze it using various classification algorithms that have been previously used to model affective states (RF, ET, LDA, LR, DT, and SVM). Our results suggest that a combination of EDA, GAZE, and TEMP data can be used to predict trust. The accuracy of the model can be further enhanced by incorporating BVP data. The RF, ET, and DT classifiers offer increased cross-validation accuracy. In the future, we plan to use the model to create a high-accuracy real-time objective trust model, which we will validate through extensive HRI studies in real-world settings. Ultimately, a robot can use this trust model in its decision making to support users effectively, while cultivating safer interactions.}

\bibliographystyle{IEEE-formatting/IEEEtran}
\bibliography{IEEE-formatting/IEEEabrv, sources}

\begin{thebibliography}{10}
\providecommand{\url}[1]{#1}
\csname url@rmstyle\endcsname
\providecommand{\newblock}{\relax}
\providecommand{\bibinfo}[2]{#2}
\providecommand\BIBentrySTDinterwordspacing{\spaceskip=0pt\relax}
\providecommand\BIBentryALTinterwordstretchfactor{4}
\providecommand\BIBentryALTinterwordspacing{\spaceskip=\fontdimen2\font plus
\BIBentryALTinterwordstretchfactor\fontdimen3\font minus \fontdimen4\font\relax}
\providecommand\BIBforeignlanguage[2]{{%
\expandafter\ifx\csname l@#1\endcsname\relax
\typeout{** WARNING: IEEEtran.bst: No hyphenation pattern has been}%
\typeout{** loaded for the language `#1'. Using the pattern for}%
\typeout{** the default language instead.}%
\else
\language=\csname l@#1\endcsname
\fi
#2}}

\bibitem{vaibhav}
V.~V. Unhelkar, S.~Li, and J.~A. Shah, ``Decision-making for bidirectional communication in sequential human-robot collaborative tasks,'' in \emph{HRI}.\hskip 1em plus 0.5em minus 0.4em\relax ACM, 2020.

\bibitem{gizem}
G.~Ate\c{s}, M.~F. St\o{}len, and E.~Kyrkjeb\o{}, ``Force and gesture-based motion control of human-robot cooperative lifting using imus,'' in \emph{HRI}.\hskip 1em plus 0.5em minus 0.4em\relax IEEE, 2022.

\bibitem{li22}
Y.~Li, C.~Lai, D.~Lala, K.~Inoue, and T.~Kawahara, ``Alzheimer's dementia detection through spontaneous dialogue with proactive robotic listeners,'' in \emph{HRI}.\hskip 1em plus 0.5em minus 0.4em\relax IEEE, 2022.

\bibitem{elgarf}
M.~Elgarf, N.~Calvo-Barajas, P.~Alves-Oliveira, G.~Perugia, G.~Castellano, C.~Peters, and A.~Paiva, ``"and then what happens?": Promoting children's verbal creativity using a robot,'' in \emph{HRI}.\hskip 1em plus 0.5em minus 0.4em\relax IEEE, 2022.

\bibitem{Gombolay2016RSS}
M.~C. Gombolay, X.~J. Yang, B.~Hayes, N.~Seo, Z.~Liu, S.~Wadhwania, T.~Yu, N.~Shah, T.~Golen, and J.~A. Shah, ``Robotic assistance in coordination of patient care.'' in \emph{RSS}, 2016.

\bibitem{mott}
T.~Mott, A.~Bejarano, and T.~Williams, ``Robot co-design can help us engage child stakeholders in ethical reflection,'' in \emph{HRI}.\hskip 1em plus 0.5em minus 0.4em\relax IEEE, 2022.

\bibitem{islam}
M.~M. Islam and T.~Iqbal, ``Hamlet: A hierarchical multimodal attention-based human activity recognition algorithm,'' in \emph{IROS}, 2020.

\bibitem{hoffman2019evaluating}
G.~Hoffman, ``Evaluating fluency in human--robot collaboration,'' \emph{IEEE Transactions on Human-Machine Systems}, 2019.

\bibitem{Iqbal2021ICRA}
T.~Iqbal and L.~D. Riek, ``{Temporal Anticipation and Adaptation Methods for Fluent Human-Robot Teaming},'' in \emph{ICRA}, 2021.

\bibitem{Islam2021_RAL}
M.~M. Islam and T.~Iqbal, ``Multi-gat: A graphical attention-based hierarchical multimodal representation learning approach for human activity recognition,'' \emph{RA-L}, 2021.

\bibitem{yasar}
M.~S. Yasar and T.~Iqbal, ``A scalable approach to predict multi-agent motion for human-robot collaboration,'' \emph{RA-L}, 2021.

\bibitem{oliveira22}
P.~Alves-Oliveira, P.~Arriaga, A.~Paiva, and G.~Hoffman, ``Children as robot designers,'' in \emph{HRI}.\hskip 1em plus 0.5em minus 0.4em\relax ACM, 2021.

\bibitem{Birmingham}
C.~Birmingham, A.~Perez, and M.~Matari\'{c}, ``Perceptions of cognitive and affective empathetic statements by socially assistive robots,'' in \emph{HRI}.\hskip 1em plus 0.5em minus 0.4em\relax IEEE, 2022.

\bibitem{hedayati}
H.~Hedayati and D.~Szafir, ``Predicting positions of people in human-robot conversational groups,'' in \emph{HRI}.\hskip 1em plus 0.5em minus 0.4em\relax IEEE, 2022.

\bibitem{mitkidis2015building}
P.~Mitkidis, J.~J. McGraw, A.~Roepstorff, and S.~Wallot, ``Building trust: Heart rate synchrony and arousal during joint action increased by public goods game,'' \emph{Physiology \& behavior}, 2015.

\bibitem{potts2019trust}
S.~R. Potts, W.~T. McCuddy, D.~Jayan, and A.~J. Porcelli, ``To trust, or not to trust? individual differences in physiological reactivity predict trust under acute stress,'' \emph{Psychoneuroendocrinology}, 2019.

\bibitem{merrill2017trust}
N.~Merrill and C.~Cheshire, ``Trust your heart: Assessing cooperation and trust with biosignals in computer-mediated interactions,'' in \emph{Proceedings of the 2017 ACM Conference on Computer Supported Cooperative Work and Social Computing}, 2017.

\bibitem{lee2004trust}
J.~D. Lee and K.~A. See, ``Trust in automation: Designing for appropriate reliance,'' \emph{Human factors}, 2004.

\bibitem{empatica}
{Empatica}, ``Empatica e4 wristband,'' \url{https://www.empatica.com/research/e4/}.

\bibitem{pupil}
{Pupil Labs}, ``Pupil invisible,'' \url{https://pupil-labs.com/products/invisible/}.

\bibitem{luhmann}
N.~Luhmann, \emph{Trust and power}.\hskip 1em plus 0.5em minus 0.4em\relax John Wiley \& Sons, 1979.

\bibitem{kramer1999trust}
R.~M. Kramer, ``Trust and distrust in organizations: Emerging perspectives, enduring questions,'' \emph{Annual review of psychology}, 1999.

\bibitem{gambetta2000can}
D.~Gambetta \emph{et~al.}, ``Can we trust trust,'' \emph{Trust: Making and breaking cooperative relations}, 2000.

\bibitem{molm2000risk}
L.~D. Molm, N.~Takahashi, and G.~Peterson, ``Risk and trust in social exchange: An experimental test of a classical proposition,'' \emph{American Journal of Sociology}, 2000.

\bibitem{dayan2010impact}
M.~Dayan and C.~A. Di~Benedetto, ``The impact of structural and contextual factors on trust formation in product development teams,'' \emph{Industrial Marketing Management}, 2010.

\bibitem{chow2008social}
W.~S. Chow and L.~S. Chan, ``Social network, social trust and shared goals in organizational knowledge sharing,'' \emph{Information \& management}, 2008.

\bibitem{fareri2012effects}
D.~S. Fareri, L.~J. Chang, and M.~R. Delgado, ``Effects of direct social experience on trust decisions and neural reward circuitry,'' \emph{Frontiers in neuroscience}, 2012.

\bibitem{lount2010impact}
R.~B. Lount~Jr, ``The impact of positive mood on trust in interpersonal and intergroup interactions.'' \emph{Personality and social psychology}, 2010.

\bibitem{mooradian2006trusts}
T.~Mooradian, B.~Renzl, and K.~Matzler, ``Who trusts? personality, trust and knowledge sharing,'' \emph{Management learning}, 2006.

\bibitem{vanneste2014trust}
B.~S. Vanneste, P.~Puranam, and T.~Kretschmer, ``Trust over time in exchange relationships: Meta-analysis and theory,'' \emph{Strategic Management Journal}, 2014.

\bibitem{fulmer2013trust}
C.~A. Fulmer and M.~J. Gelfand, ``How do i trust thee? dynamic trust patterns and their individual and social contextual determinants,'' \emph{Models for intercultural collaboration and negotiation}, 2013.

\bibitem{jones1998experience}
G.~R. Jones and J.~M. George, ``The experience and evolution of trust: Implications for cooperation and teamwork,'' \emph{Academy of management review}, 1998.

\bibitem{kim2009repair}
P.~H. Kim, K.~T. Dirks, and C.~D. Cooper, ``The repair of trust: A dynamic bilateral perspective and multilevel conceptualization,'' \emph{Academy of Management Review}, 2009.

\bibitem{herse}
S.~Herse, J.~Vitale, B.~Johnston, and M.-A. Williams, ``Using trust to determine user decision making and task outcome during a human-agent collaborative task,'' in \emph{HRI}.\hskip 1em plus 0.5em minus 0.4em\relax ACM, 2021.

\bibitem{xu}
A.~Xu and G.~Dudek, ``Optimo: Online probabilistic trust inference model for asymmetric human-robot collaborations,'' in \emph{HRI}, 2015.

\bibitem{hancock}
P.~A. Hancock, D.~R. Billings, K.~E. Schaefer, J.~Y.~C. Chen, E.~J. de~Visser, and R.~Parasuraman, ``A meta-analysis of factors affecting trust in human-robot interaction,'' \emph{Human Factors}, 2011.

\bibitem{kok2020trust}
B.~C. Kok and H.~Soh, ``Trust in robots: Challenges and opportunities,'' \emph{Current Robotics Reports}, 2020.

\bibitem{azevedo2021unified}
H.~Azevedo-Sa, X.~J. Yang, L.~P. Robert, and D.~M. Tilbury, ``A unified bi-directional model for natural and artificial trust in human-robot collaboration,'' \emph{RA-L}, 2021.

\bibitem{chen2020trust}
M.~Chen, S.~Nikolaidis, H.~Soh, D.~Hsu, and S.~Srinivasa, ``Trust-aware decision making for human-robot collaboration: Model learning and planning,'' \emph{THRI}, 2020.

\bibitem{bryant}
D.~Bryant, J.~Borenstein, and A.~Howard, ``Why should we gender? the effect of robot gendering and occupational stereotypes on human trust and perceived competency,'' in \emph{HRI}.\hskip 1em plus 0.5em minus 0.4em\relax ACM, 2020.

\bibitem{greenHRI}
H.~N. Green, M.~M. Islam, S.~Ali, and T.~Iqbal, ``Who's laughing nao? examining perceptions of failure in a humorous robot partner,'' in \emph{HRI}, 2022.

\bibitem{robinette2017}
P.~Robinette, A.~M. Howard, and A.~R. Wagner, ``Effect of robot performance on human–robot trust in time-critical situations,'' \emph{IEEE Transactions on Human-Machine Systems}, 2017.

\bibitem{natarajan}
M.~Natarajan and M.~Gombolay, \emph{Effects of Anthropomorphism and Accountability on Trust in Human Robot Interaction}.\hskip 1em plus 0.5em minus 0.4em\relax ACM, 2020.

\bibitem{green}
H.~N. Green, M.~M. Islam, S.~Ali, and T.~Iqbal, ``ispy a humorous robot: Evaluating the perceptions of humor types in a robot partner,'' in \emph{AAAI Putting AI in the Critical Loop: Assured Trust and Autonomy in Human-Machine Teams}, 2022.

\bibitem{robinette}
P.~Robinette, W.~Li, R.~Allen, A.~M. Howard, and A.~R. Wagner, ``Overtrust of robots in emergency evacuation scenarios,'' in \emph{HRI}, 2016.

\bibitem{han}
Z.~Han, E.~Phillips, and H.~A. Yanco, ``The need for verbal robot explanations and how people would like a robot to explain itself,'' \emph{HRI}, 2021.

\bibitem{williams}
T.~Williams, P.~Briggs, and M.~Scheutz, ``Covert robot-robot communication: Human perceptions and implications for human-robot interaction,'' \emph{HRI}, 2015.

\bibitem{martelaro}
N.~Martelaro, V.~C. Nneji, W.~Ju, and P.~Hinds, ``Tell me more designing hri to encourage more trust, disclosure, and companionship,'' in \emph{HRI}, 2016.

\bibitem{clair}
A.~S. Clair and M.~Matarić, ``How robot verbal feedback can improve team performance in human-robot task collaborations,'' in \emph{HRI}, 2015.

\bibitem{schule}
M.~Sch\"{u}le, J.~M. Kraus, F.~Babel, and N.~Rei\ss{}ner, ``Patients' trust in hospital transport robots: Evaluation of the role of user dispositions, anxiety, and robot characteristics,'' in \emph{HRI}.\hskip 1em plus 0.5em minus 0.4em\relax IEEE, 2022.

\bibitem{myers2016influence}
C.~D. Myers and D.~Tingley, ``The influence of emotion on trust,'' \emph{Political Analysis}, 2016.

\bibitem{dunn2005feeling}
J.~R. Dunn and M.~E. Schweitzer, ``Feeling and believing: the influence of emotion on trust.'' \emph{Personality and social psychology}, 2005.

\bibitem{chu2024physiology}
B.~Chu, K.~Marwaha, T.~Sanvictores, A.~O. Awosika, and D.~Ayers, ``Physiology, stress reaction,'' in \emph{StatPearls [Internet]}.\hskip 1em plus 0.5em minus 0.4em\relax StatPearls Publishing, 2024.

\bibitem{kim2018stress}
H.-G. Kim, E.-J. Cheon, D.-S. Bai, Y.~H. Lee, and B.-H. Koo, ``Stress and heart rate variability: a meta-analysis and review of the literature,'' \emph{Psychiatry investigation}, 2018.

\bibitem{herborn2015skin}
K.~A. Herborn, J.~L. Graves, P.~Jerem, N.~P. Evans, R.~Nager, D.~J. McCafferty, and D.~E. McKeegan, ``Skin temperature reveals the intensity of acute stress,'' \emph{Physiology \& behavior}, 2015.

\bibitem{reinhardt2012salivary}
T.~Reinhardt, C.~Schmahl, S.~W{\"u}st, and M.~Bohus, ``Salivary cortisol, heart rate, electrodermal activity and subjective stress responses to the mannheim multicomponent stress test (mmst),'' \emph{Psychiatry research}, 2012.

\bibitem{akalin2022you}
N.~Akalin, A.~Kristoffersson, and A.~Loutfi, ``Do you feel safe with your robot? factors influencing perceived safety in human-robot interaction based on subjective and objective measures,'' \emph{International journal of human-computer studies}, 2022.

\bibitem{ahmad2023crucial}
M.~Ahmad and A.~Alzahrani, ``Crucial clues: Investigating psychophysiological behaviors for measuring trust in human-robot interaction,'' in \emph{Proceedings of the 25th International Conference on Multimodal Interaction}, 2023.

\bibitem{bailey2007automation}
N.~R. Bailey and M.~W. Scerbo, ``Automation-induced complacency for monitoring highly reliable systems: the role of task complexity, system experience, and operator trust,'' \emph{Theoretical Issues in Ergonomics Science}, 2007.

\bibitem{zeng2006spontaneous}
Z.~Zeng, Y.~Fu, G.~I. Roisman, Z.~Wen, Y.~Hu, and T.~S. Huang, ``Spontaneous emotional facial expression detection.'' \emph{J. Multim.}, 2006.

\bibitem{panda}
{Franka Emika}, ``Franka research 3,'' \url{https://www.franka.de/research}.

\bibitem{ekman1978facial}
P.~Ekman and W.~V. Friesen, ``Facial action coding system,'' \emph{Environmental Psychology \& Nonverbal Behavior}, 1978.

\bibitem{muir1996trust}
B.~M. Muir and N.~Moray, ``Trust in automation. part ii. experimental studies of trust and human intervention in a process control simulation,'' \emph{Ergonomics}, 1996.

\bibitem{desai}
M.~Desai, P.~Kaniarasu, M.~Medvedev, A.~Steinfeld, and H.~Yanco, ``Impact of robot failures and feedback on real-time trust,'' in \emph{HRI}, 2013.

\bibitem{chen2018planning}
M.~Chen, S.~Nikolaidis, H.~Soh, D.~Hsu, and S.~Srinivasa, ``Planning with trust for human-robot collaboration,'' in \emph{HRI}, 2018.

\bibitem{washburn2020robot}
A.~Washburn, A.~Adeleye, T.~An, and L.~D. Riek, ``Robot errors in proximate hri: how functionality framing affects perceived reliability and trust,'' \emph{ACM THRI}, 2020.

\bibitem{openface2}
T.~Baltrusaitis, A.~Zadeh, Y.~C. Lim, and L.-P. Morency, ``Openface 2.0: Facial behavior analysis toolkit,'' in \emph{2018 13th IEEE International Conference on Automatic Face \& Gesture Recognition (FG 2018)}, 2018.

\bibitem{boucsein2012electrodermal}
W.~Boucsein, \emph{Electrodermal activity}.\hskip 1em plus 0.5em minus 0.4em\relax Springer Science \& Business Media, 2012.

\bibitem{wilke2007short}
K.~Wilke, A.~Martin, L.~Terstegen, and S.~Biel, ``A short history of sweat gland biology,'' \emph{International journal of cosmetic science}, 2007.

\bibitem{lui2022apple}
G.~Y. Lui, D.~Loughnane, C.~Polley, T.~Jayarathna, and P.~P. Breen, ``The apple watch for monitoring mental health--related physiological symptoms: Literature review,'' \emph{JMIR Mental Health}, 2022.

\bibitem{gan2016overview}
S.~K.-E. Gan, C.~Koshy, P.-V. Nguyen, and Y.-X. Haw, ``An overview of clinically and healthcare related apps in google and apple app stores: connecting patients, drugs, and clinicians,'' \emph{Scientific phone apps and mobile devices}, 2016.

\bibitem{evans2017myfitnesspal}
D.~Evans, ``Myfitnesspal,'' \emph{British Journal of Sports Medicine}, 2017.

\end{thebibliography}

\end{document}